# Cleaning Schedule Optimization of Heat Exchanger Networks Using Particle Swarm Optimization

**Totok R. Biyanto, Sumitra Wira Suganda, Matraji, Yerry Susatio, Heri Justiono, Sarwono**
Jurusan Teknik Fisika FTI-ITS Surabaya
Keputih Sukolilo, Surabaya 60111, Indonesia
trb@epits.ac.id

**ABSTRACT**

Oil refinery is one of industries that require huge energy consumption. The today technology advance requires energy saving. Heat integration is a method used to minimize the energy comsumption though the implementation of Heat Exchanger Network (HEN). CPT is one of types of Heat Exchanger Network (HEN) that functions to recover the heat in the flow of product or waste. HEN comprises a number of heat exchangers (HEs) that are serially connected. However, the presence of fouling in the heat exchanger has caused the decline of the performance of both heat exchangers and all heat exchanger networks. Fouling can not be avoided. However, it can be mitigated. In industry, periodic heat exchanger cleaning is the most effective and widely used mitigation technique. On the other side, a very frequent cleaning of heat exchanger can be much costly in maintenance and lost of production. In this way, an accurate optimization technique of cleaning schedule interval of heat exchanger is very essential. Commonly, this technique involves three elements: model to simulate the heat exchanger network, representative fouling model to describe the fouling behavior and suitable optimization algorithm to solve the problem of clening schedule interval for heat exchanger network. This paper describe the optimization of interval cleaning schedule of HEN within the 44-month period using PSO (particle swarm optimization). The number of iteration used to achieve the convergent is 100 iterations and the fitness value in PSO correlated with the amount of heat recovery, cleaning cost, and additional pumping cost. The saving after the optimization of cleaning schedule of HEN in this research achieved at $ 1.236 millions or 23% of maximum potential savings.

## 1 INTRODUCTION

Oil refinery and general industry today have been trying to minimize the energy consumption. To do so, heat integration becomes one of interesting methods. It uses the heat from product or waste as the energy source to heat up the cold flow. One of technologies often used for heat integration in oil industry is Crude Preheat Train (CPT) (Macchietto S, 2009).

Heat exchanger networks are formed by connecting a number of heat exchangers in series and/or parallel configurations. The presence of fouling in HE will reduce the HEN performance. Fouling is a formation of deposit layers in heat exchanger that can impact on the reduce of the heat exchanger (HE) performance (Pogiatzis, et al., 2011). The pressure drop across the fouled heat exchanger units increases due to the reduction in the flow area and consequently increases the pumping costs.

In term of many problems caused by fouling, Engineering Sciences Data Unit (ESDU) affirms that fouling in the heat exchanger network is a very serious problem (ESDU, 2000). This problem is related to the increase of energy consumption, economical loses, increase of carbon dioxide ($CO_2$) and additional pressure drop (Yeap, et al., 2005). Fouling is unavoidable, but it can be mitigated through effective implementation of appropriate fouling mitigation techniques.

Fouling mitigation techniques include addition of antifoulant chemicals to the refinery feed, design and use of more efficient heat exchangers (Samaili, 2001) and periodic cleaning of heat exchangers (Pogiatzis, et al., 2011). Besides their capability in mitigating the fouling, each technique has its own drawbacks

Periodic cleaning of heat exchangers is an option to overcome the losses due to fouling. However, cleaning activities in HENs are cost intensive. Materials and tools required for cleaning, labor costs, shutting down the heat exchangers during cleaning, and disposal of cleaning wastes are examples of the cost attributed for the cleaning activities. Therefore, an optimization of cleaning schedule for a given HEN is important in order to realize the potentials of heat integration.

One of journals written by Smaili, et al. (2001) explained that optimization of HEN schedule involves three elements:

1. Model that can simulate HEN.
2. Fouling model to describe the fouling behavior
3. Optimization method for HEN cleaning schedule, Mixed Integer NonLinear Programing (MINP) class. (Biyanto, 2013).

This paper describe the HEN simulation method was performed using mass and energy balance equation (Ishiyama, et al, 2010), empirical fouling model (Sanaye, et al., 2007) and particle swarm optimization (PSO) to optimize the interval of HEN cleaning schedule.

## 2 HEAT EXCHANGER AND HEAT EXCHANGER NETWORK (HEN)

Traditionally, the heat exchanger performance analysis and simulation are performed using steady-state energy balance across the heat exchanger. The energy balance on the hot and cold fluids together with the heat-transfer equation constitutes the model of heat exchangers. A simplified model generally uses an average driving force such as log mean temperature difference (LMTD) and assumes uniform properties of the fluids along the length of the heat exchanger to determine the overall heat-transfer coefficient.

Under the assumption that there is no heat loss to the surroundings, the heat lost by the hot fluid stream shall be equal to the heat gained by the cold fluid stream, thus

$$Q_c = Q_h \tag{1}$$

where $Q_c$ = amount of heat received by cold fluid
$Q_h$ =amount of heat released by hot fluid

The amount of heat received by the cold fluid, $Q_c$, is given by

$$Q_c = m_c c_{p,c}\left(T_{c,o} - T_{c,i}\right) \tag{2}$$

where $m_c$ = mass flow rate of the cold fluid (crude oil)
$c_{p,c}$ = specific heat of the cold fluid
$T_{c,i}$ = inlet temperature of the cold fluid
$T_{c,o}$ = outlet temperature of the cold fluid

The amount of heat released (lost) by the hot fluid, $Q_h$, is given by

$$Q_h = m_h c_{p,h}\left(T_{h,i} - T_{h,o}\right) \tag{3}$$

where $m_h$ = mass flow rate of the hot fluid

$c_{p,h}$ = specific heat of the hot fluid
$T_{h,i}$ = inlet temperature of the hot fluid
$T_{h,o}$ = outlet temperature of the hot fluid

The amount of heat transferred from the hot fluid to the cold fluid, $Q$, across the heat exchanger surface would be equal to $Q_c$ and $Q_h$ and is given by

$$Q = UAF\Delta T_{lm} \tag{4}$$

where $U$ = overall heat-transfer coefficient
$A$ = heat-transfer surface area
$\Delta T_{lm}$ = Log Mean Temperature Difference (LMTD)
$F$ = LMTD correction factor.

The overall heat-transfer coefficient, $U$, is determined using empirical correlations of the individual film heat-transfer coefficients and the resistance due to fouling as given by

$$\frac{1}{U} = \frac{d_o}{d_i h_i} + \frac{d_o R_{f,i}}{d_i} + \frac{d_o \ln\left(\frac{d_o}{d_i}\right)}{2k_w} + R_{f,o} + \frac{1}{h_o} \tag{5}$$

where: $R_{f,i}$ = inside fouling resistance
$R_{f,o}$ = outside fouling resistance
$h_i$ = tube-side film heat-transfer coefficient
$h_o$ = shell-side film heat-transfer coefficient
$U$ = overall heat-transfer coefficient
$k_w$ = thermal conductivity of the tube metal
$d_o$ = outside diameter of the tube
$d_i$ = inside diameter of the tube

The heat exchanger model equations described above constitutes the simulation model of a heat exchanger. Steady state solutions of the heat exchanger models determine outlet temperatures of hot and cold streams. Since the crude oil flows from one heat exchanger to the next and some of the heating mediums flow through more than one heat exchanger in series, the heat exchanger models cannot be solved independently. A simultaneous solution is required to obtain the temperature of outlet streams in all heat exchangers in the CPT. Since heat-transfer rate from hot to cold fluid will be equal to the change of enthalpy of the hot fluid, the outlet temperature of cold and hot fluids in a heat exchanger

$$T_{c,o} = \left[\frac{k_1\left(\exp\left(-k_2 F\left(k_1 - 1\right)\right) - 1\right)}{\exp\left(-k_2 F\left(k_1 - 1\right)\right) - k_1}\right] T_{h,i} + \left[\frac{\left(1 - k_1\right)\exp\left(-k_2 F\left(k_1 - 1\right)\right)}{\exp\left(-k_2 F\left(k_1 - 1\right)\right) - k_1}\right] T_{c,i} \tag{6}$$

$$T_{h,o} = \left[\frac{\exp\left(-k_2 F\left(k_1 - 1\right)\right) - 1}{\exp\left(-k_2 F\left(k_1 - 1\right)\right) - k_1}\right] T_{c,i} + \left[\frac{\left(k_1 - 1\right)}{\exp\left(-k_2 F\left(k_1 - 1\right)\right) - k_1}\right] T_{h,i} \tag{7}$$

where,

$$k_1 = \frac{m_h c_{p,h}}{m_c c_{p,c}}$$

$$k_2 = \frac{UA}{m_h c_{p,h}}$$

Equations (6) and (7) apply to each heat exchanger in the network.

Figure 1 shows the heat exchanger networks under study. HEN are formed by connecting a number of heat exchangers in series and/or parallel configurations. Figure 1 show the structure of HEN under study that consists of 11 heat exchangers.

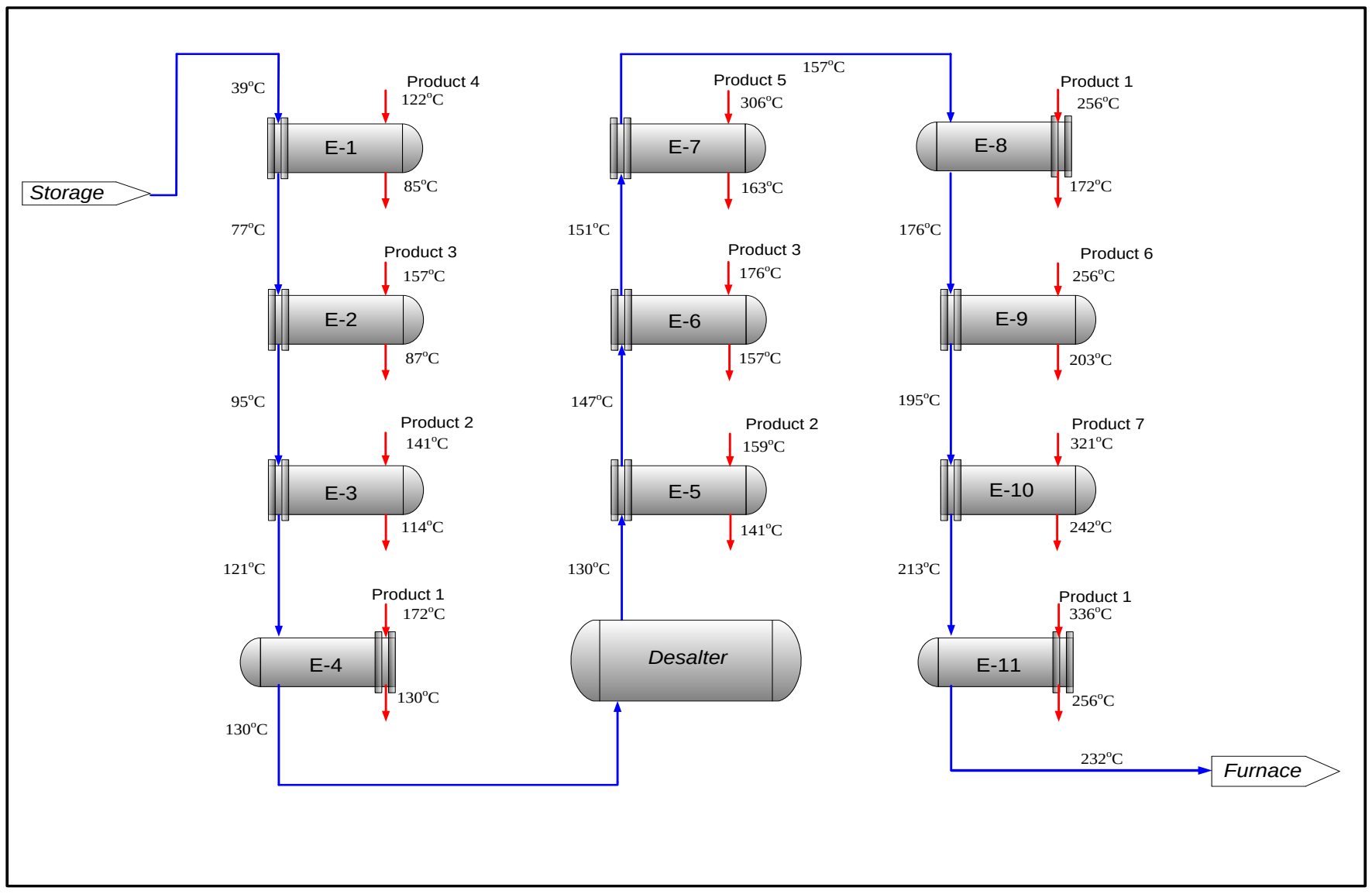

Figure 1: Scheme of Heat Exchanger Network (HEN)

## 3 FOULING MODEL

One of the simplest models to describe the fouling behaviour was put forward by Kern and Seaton (1959). Basically, this model is a mathematical interpretation of asymptotic model. The mathematical formula of this model is given by

$$R_f(t) = a(1 - e^{-bt}) \tag{8}$$

## 4 PROBLEM FORMULATION

The heat exchangers in the CPT need to be cleaned periodically in order to reduce the unrecovered heat loss and the additional pressure drop due to fouling. However, the heat exchanger cleaning is considerably expensive and it causes disruptions in the plant production. Higher cost of cleaning would be incurred when the heat exchangers are cleaned too frequently while less frequent cleanings lead to higher operating costs due to increased heat loss. Therefore, an optimal cleaning schedule that provides maximum economic savings with minimum energy losses is required to be established. In general, the cost function of the cleaning schedule optimization problem can be stated as (Biyanto, 2013):

Minimize total cost = cost of energy loss +
cost of cleaning the heat exchangers +

cost of additional pumping power

$$J = \sum_{n=1}^{N_E}\sum_{t=1}^{t_F}\left[C_E\left(Q_{n,t}^{i} - Q_{n,t}^{a}\right)y_{n,t} + C_{cl,n}(1-y_{n,t}) + C_p\left(W_{P,n,t}\right)y_{n,t}\right] \tag{9}$$

subject to :

$$Q_{n,t}^{a} = f_1(m_{c/h}, \rho_{c/h}, \mu_{c/h}, c_{p,c/h}, k_{c/h}, T_{i,c/h}, U_a, n, t)$$

$$Q_{n,t}^{i} = f_2(m_{c/h}, \rho_{c/h}, \mu_{c/h}, c_{p,c/h}, k_{c/h}, T_{i,c/h}, U_i, n, t)$$

where the functions $f_1$ and $f_2$ are the heat exchanger model equations under actual and ideal conditions, respectively. $C_{cl}$ is cleaning cost, $C_E$ is the unit energy cost, $N_E$ is the number of heat exchangers, $W_P$. Additional pumping cost incorporates cost of pump work, $C_p$, or electricity unit cost that is assumed constant, The binary variable, $y_{n,t}$, indicates the status of the heat exchanger. A value of $y_{n,t}$ = 1 indicates that the heat exchanger '*n*' is in operation on day '*t*' and a value of $y_{n,t}$ = 0 means the heat exchanger undergoes cleaning.

## 5 PARTICLE SWARM OPTIMIZATION (PSO)

Particle swarm optimization (PSO) is an evolutionary computation technique that was developed by Kennedy and Eberhart in 1995. The concept of PSO is a group of particles forming a social population or frequently described with a group of birds in a social population (Kennedy J, et al., 2007). PSO can be easily implemented and it is computationally inexpensive (Parsopoulus K E, et al., 2002). The steps in making algorithm of optimization of schedule interval for heat exchanger cleaning using PSO are presented as follows:

1. Randomly generating the particles' position with dimension parameter for 11 heat exchangers and initial speed. The number of particles used to seek the value of optimization of schedule interval for HE cleaning and the dimension in PSO is to represent the cleaning period of each heat exchanger. The initial position and speed are randomly determined with the limitation for the initial position ranging from 0 to 31 and limitation for initial speed ranging from 0 to 1.
2. Determining the value of inertial weight (θ) for each of iteration in PSO using Eq. (10).

$$\theta(i) = \theta_{max} - e^{\left(\frac{\theta_{max}-\theta_{min}}{\theta_{min}}\right)i} \tag{10}$$

   where:
   $i$ = iteration 1,2,3,4...
   $\theta_{max}$ = initial value of inertial weight
   $\theta_{min}$ = final value of inertial weight
3. Determining the cleaning schedule ranging in 44 months.
4. Evaluating the fitness value or Objective Function from each particle based on its position. The fitness value refers to the amount of cost for energy recovery, cleaning, and pumping. The fitness value is a formulation of objective function in Eq. (9).
   Explanation:
5. Determining the best initial position for each particle "Pbest" with the lowest value of objective function (minimum) and the best initial position for all particles "Gbest" with the lowest value of objective function from all particles (minimum).
6. In the iteration i, determining some variables to achieve the optimum value.

a. To update the newest speed from each particle using Eq. (11).

$$V_{id} = V_{id} + c_1 \times rand() \times (p_{id} + x_{id}) + c_2 \times rand() \times (p_{gd} x_{id}) \tag{11}$$

b. To update the newest position of each particle using Eq. (12).

$$X_{id} = X_{id} + V_{id} \tag{12}$$

c. To evaluate the fitness value from the newest position and speed for each particle. This evaluation uses Eq.(9).
d. To update the best position from the particle itself stated in 'pbest' and the best position form all particles stated in "gbest".
e. To update the cleaning schedule for each heater exchanger using the best position from all particles in each dimension.

7. Repeating the step 3 if the iteration has not reached 100.

## 6 RESULT AND DISCUSSION

The optimization of cleaning schedule for HEN was conducted within the period of 44 months. The convergent condition in Figure 2 was obtained when the fitness value is about 1.32 x $10^7$ at iteration 54. The best position of all particles (Gbest) from the result of cleaning schedule for HEN is represented in Table 1. The Gbest value was used as the cleaning interval of each exchanger within 44 months. The use of 11 coulumn was performed to represent the cleaning schedule of each heat exchanger. Coulmn 1 represented cleaning schedule of E-01 until column 11 representing the cleaning schedule of E-11. The number of cleaning schedule during period 44 months of each HE from the result of optimization used PSO as presented in Table 2.

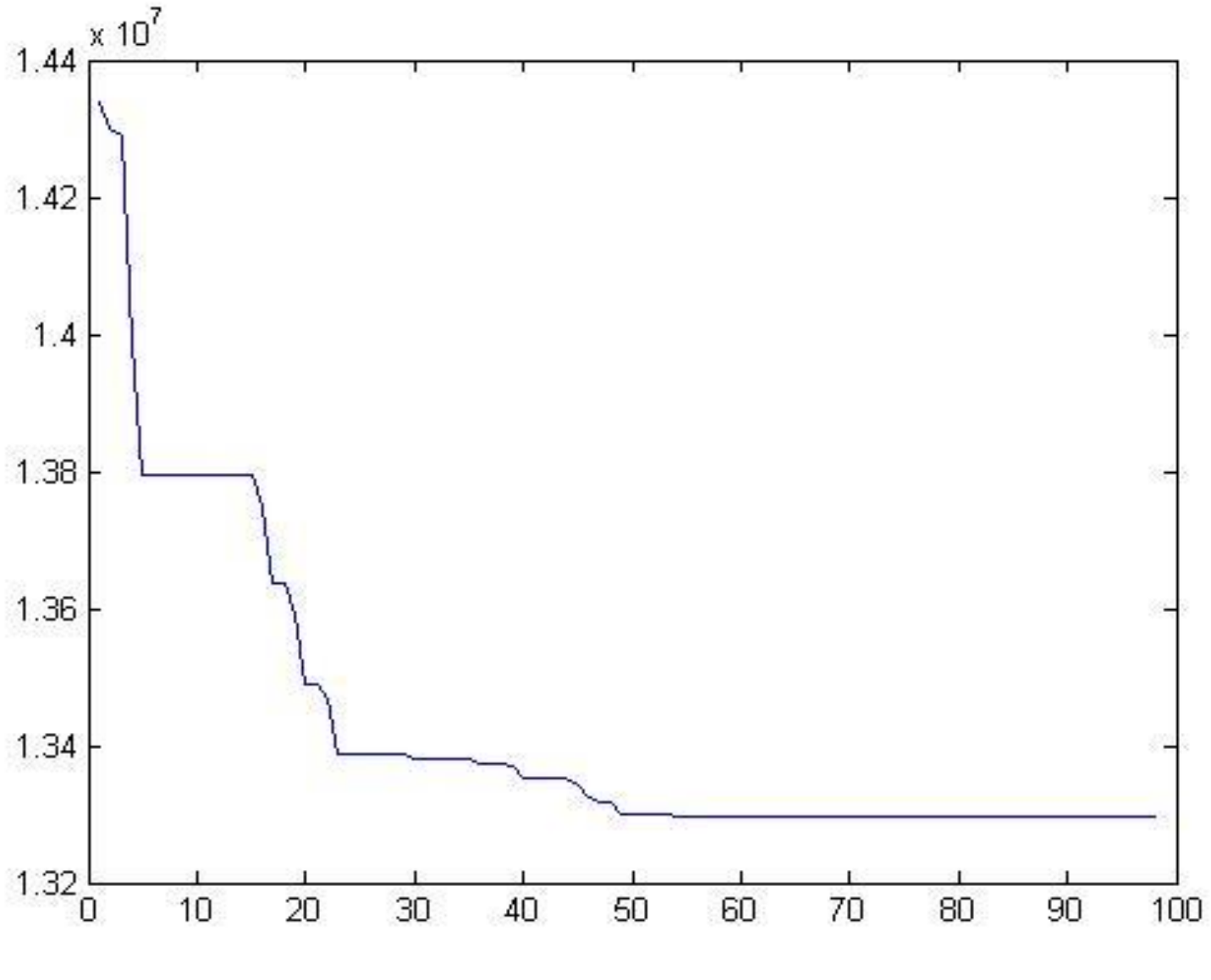

Figure 2: Graphic of fitness value (minimum) from the objective function of optimization

Table 1: The best position of all particles (Gbest) or cleaning schedule interval

| E-1 | E-2 | E-3 | E-4 | E-5 | E-6 | E-7 | E-8 | E-9 | E-10 | E-11 |
|---|---|---|---|---|---|---|---|---|---|---|
| 16 | 23 | 28 | 9 | 5 | 9 | 28 | 5 | 9 | 5 | 24 |

Table 2: Numer of cleaning schedule for each HE during 44 months

| E-1 | E-2 | E-3 | E-4 | E-5 | E-6 | E-7 | E-8 | E-9 | E-10 | E-11 |
|---|---|---|---|---|---|---|---|---|---|---|
| 2 | 1 | 1 | 4 | 8 | 4 | 1 | 8 | 4 | 8 | 1 |

The result of optimization of cleaning schedule as given in Table 1 and Table 2 was influenced by several factors including fouling resistance, additional pumping cost, cleaning cost and recovered heat for each heat exchanger, as shown in Eqs. (1) – (9).

The detailed costs of the individual terms in the cost function and the net loss corresponding to the optimal cleaning schedule is shown in Table 3. It is observed that a net savings of US$ 1.236 million or 23% of maximum potential savings was achieved by implementing the optimal cleaning schedule. The net loss in energy recovery has decreased from RM 18.57 million under the conditions of no cleaning to RM 21.14 million.

Table 3: The detail cost of energy recovery, cleaning cost, and pumping cost

| Condition | Recovered Energy ($) | Cleaning cost ($) | Pumping cost ($) |
|---|---|---|---|
| **Clean** | 23.437.800 | 0 | 236.909 |
| **Fouled** | 18.570.433 | 0 | 596.365 |
| **Cleaning Schedule** | 21.138.727 | 1.478.400 | 449.504 |

It is observed that only the heat exchangers E-1, E-5, E-6 and E-8 provided net savings in energy recovery. All other heat exchangers have resulted in additional losses under the conditions of cleaning schedule conditions. However, their indirect effect on the performance of other heat exchangers have resulted in a net savings of US$ 1.236 million. It proves that HEN cleaning schedule optimization is a hard problem to solve. The energy recoveries under clean, fouled and cleaning schedule conditions are shown in Figure 3.

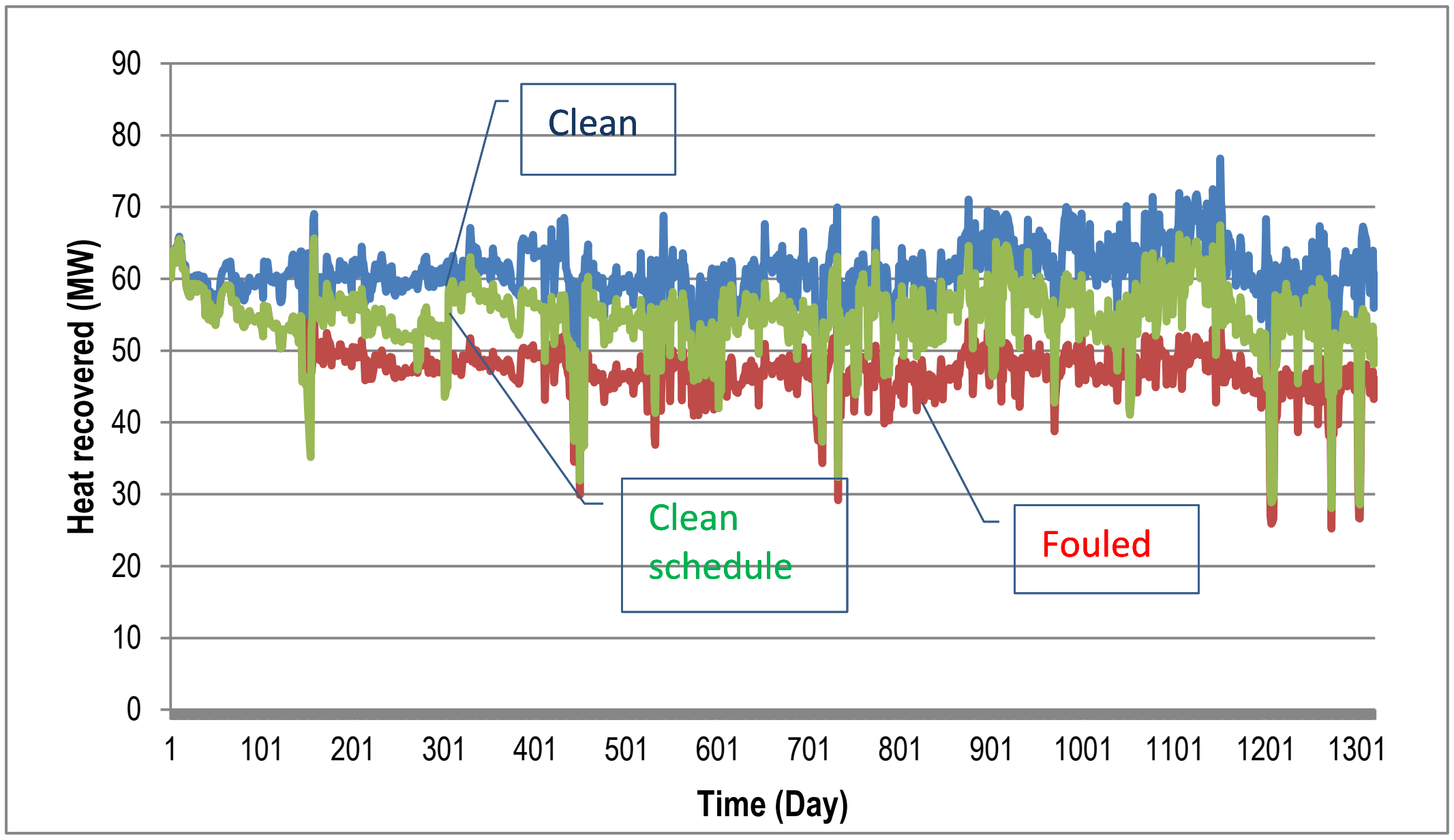

Figure 3: Heat duty under fouled, clean and cleaning schedule condition

A realistic cleaning schedule optimization problem was formulated which includes the additional pumping cost in the cost function. The number of optimization variables were considerably reduced through the use of cleaning interval for each heat exchanger. By nature of the optimization problem, which is an MINLP problem, was solved using simple PSO. The use of PSO to solve the optimization problem did not involve any approximations or assumptions to simplify the problem.

## 7 CONCLUSION

The conclusions of this research can be drawn as follow:

1. The saving resulted during 44 months after optimization was at US$ 1.236 million or 23% of maximum potential savings
2. It is observed that only the heat exchangers E-01, E-05, E-06 and E-8 provided net savings in energy recovery and the remaining heat exchangers experiencing further loss.
3. The use of PSO to solve the optimization problem did not involve any approximations or assumptions to simplify the problem.